\documentclass{article}

\PassOptionsToPackage{sort&compress}{natbib}
\usepackage{iclr2027_conference,times}
\setcitestyle{numbers,square,citesep={,}}

\usepackage{graphicx}
\usepackage{wrapfig}
\usepackage{booktabs}
\usepackage{amsmath,amssymb}
\usepackage{multirow}
\usepackage{xcolor}
\usepackage{hyperref}
\usepackage{url}
\hypersetup{hidelinks}

\title{SCoPE: Training-Free Audio-Visual Event Perception via Sparse Cross-Modal Prior Exchange}

\author{
\makebox[\dimexpr\textwidth-2\tabcolsep-4pt\relax][c]{%
\shortstack{
Jaemo Jeong \quad Junho Yoon \quad Hyunju Kim \quad Dongman Lee\\
KAIST\\
\texttt{\{gosfl4760, vpdtlrdl, iplay93, dlee\}@kaist.ac.kr}
}}
}

\iclrfinalcopy

\newcounter{lemma}
\newcounter{definition}

\begin{document}

\maketitle
\lhead{}
\renewcommand{\headrulewidth}{0pt}

\begin{abstract}
Audio-visual event perception (AVEP) determines which events occur in a video, when they occur, and whether they are audible, visible, or both. Training-free methods query new event vocabularies by matching frozen audio and visual features with text-encoded event names. However, related labels share evidence. An incorrect label can then score at least as high as a correct one. We call this a false co-activation (FCA). No scalar cutoff can reject the incorrect label while keeping every correct one. Class-specific thresholds may prevent that label from becoming a final prediction, but the FCA remains in the underlying score vector. We introduce SCoPE, a training-free framework in which all queried labels compete for shared evidence and each modality guides event selection in the other. We derive an exact condition for when this competition removes an FCA in a two-label fit. With identical frozen CLIP+CLAP backbones on LLP, SCoPE improves Type@seg by 7.45 points and Event@seg by 5.04 points compared with the reported AV$^2$A values. The same fixed configuration transfers unchanged to OV-AVEBench and VGGSound-AVEL100k.
\end{abstract}

\section{Introduction}

Audio-visual event perception (AVEP)~\cite{tian2020avvp,gao2023cmpae,zhou2024label} determines which events occur in a video, when they occur, and whether each is audible, visible, or present in both modalities. This distinction matters for robotics and autonomous driving~\cite{chen2021semanticavnav,gan2019stereosound} when a source is outside the camera view, visually occluded, or acoustically masked.

Most AVEP systems learn a fixed event vocabulary. Supporting even one new class means collecting labeled videos and retraining the model. Recent training-free methods~\cite{zhou2025ovavel,shaar2025av2a,cho2026cue2rule} instead compare frozen audio and visual features with text-encoded event names. Each event name is matched to every time interval separately, and the resulting decisions are adjusted through fixed or adaptive score thresholds.

However, related event names lie close in the embedding space, so the same evidence can raise several labels. An incorrect label can then score at least as high as a correct one. We call this a \emph{false co-activation} (FCA). No scalar cutoff can reject the incorrect label while keeping every correct one. Class-specific thresholds may prevent that label from becoming a final prediction, but the FCA remains in the independently matched scores. Section~\ref{sec:fca_analysis} gives the exact definition.

We divide FCAs into two types. The first occurs when an incorrect label is sandwiched between a stronger and a weaker correct label in the score ranking. In Figure~\ref{fig:motivation}(a), the video contains \textit{Dog} and \textit{Car}, but not \textit{Cat}. Yet \textit{Cat} responds to the visual evidence for \textit{Dog} and ranks above \textit{Car}. Its score is largely borrowed from the stronger correct label, so we call this case a \emph{shadow}. The second occurs when the available evidence does not clearly distinguish the incorrect label from a correct one. In the audio branch, a car engine can also sound like a motorcycle, so \textit{Motorcycle} remains plausible beside \textit{Car}. We call this case a \emph{rival}. Section~\ref{sec:fca_analysis} defines both types exactly.

\begin{center}
\begin{minipage}{0.92\linewidth}
    \centering
    \includegraphics[width=\linewidth]{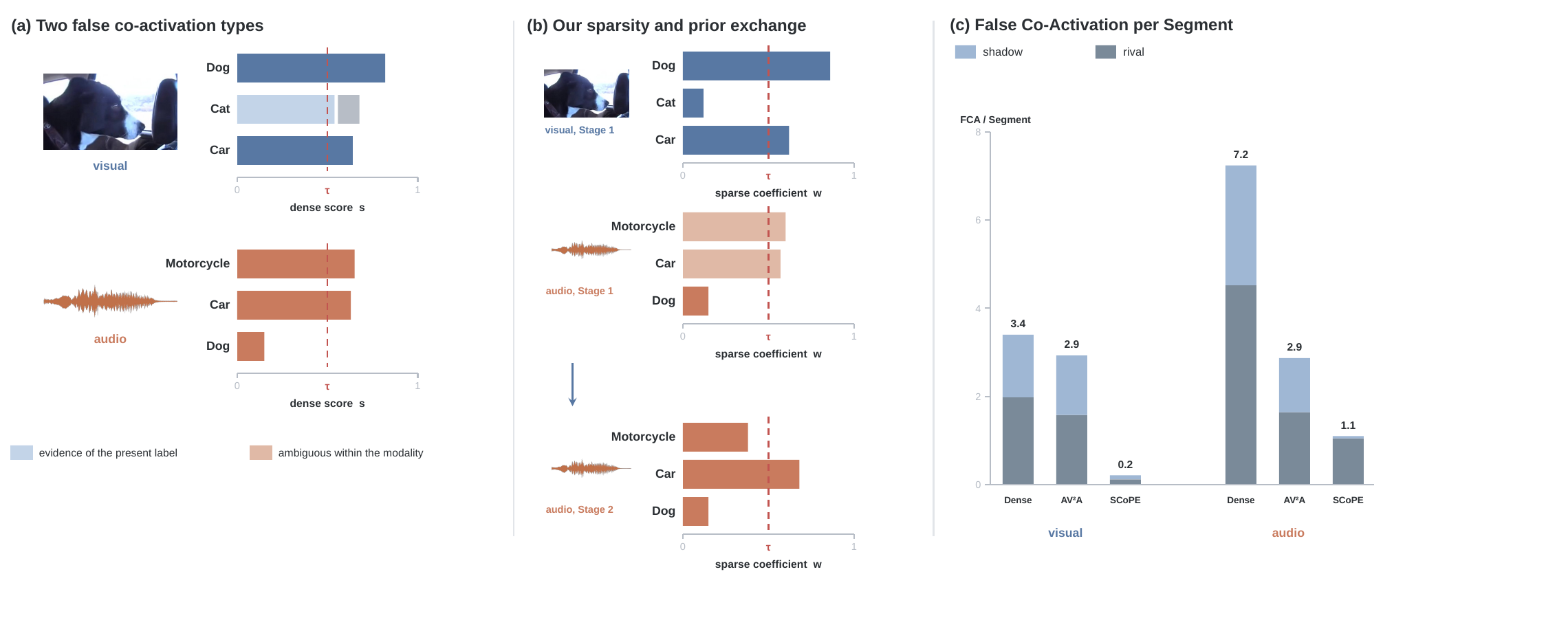}
    \par\smallskip
    \refstepcounter{figure}
    \label{fig:motivation}
    \parbox{\linewidth}{\small Figure~\thefigure: False co-activation and SCoPE. (a) No scalar cutoff keeps both annotated events while rejecting the extra candidate between them. In a two-label sparse fit, \textit{Dog} remains while \textit{Cat} drops out. \textit{Motorcycle} remains beside \textit{Car}. (b) Stage~1 selects sparse support, and Stage~2 uses cross-modal support to change the target's per-class $\ell_1$ selection costs before re-solving. (c) Mean FCA counts on multi-label LLP segments, split by the two pairwise diagnostic tags.}
\end{minipage}
\end{center}

We propose \textbf{SCoPE}, a training-free framework for \textbf{S}parse \textbf{C}ross-modal \textbf{P}rior \textbf{E}xchange. Stage~1 makes all labels compete for the evidence in each time interval, reducing FCAs before cross-modal guidance. Stage~2 uses events selected in one modality to lower the cost of selecting the same events in the other. The target modality then makes its own decision from its own features rather than copying the source score or prediction. A standard sparse readout cuts the sorted positive coefficients at their largest drop and therefore needs no calibrated cutoff. SCoPE requires no task-specific training, instruction tuning, or LLM decoding.

Our contributions are:
\begin{itemize}
    \item We define score-level false co-activation and show why one scalar cutoff on a fixed score vector cannot reject an incorrect label without losing a correct one. We derive an exact condition for the two-label fit and test its prediction with the full 25-label dictionary.
    \item We introduce sparse cross-modal prior exchange. Events selected in one modality lower the cost of selecting the same events in the other. Each modality still makes its own prediction instead of copying source scores or labels.
    \item Compared with the reported AV$^2$A values obtained with frozen CLIP+CLAP encoders, SCoPE improves Type@seg by 7.45 points and Event@seg by 5.04 points on LLP.
    \item SCoPE's sparse inference precedes final decision rules, so it can be combined with existing training-free post-processing. Adding AV$^2$A-style dynamic thresholds and candidate filtering raises Type@seg from 51.25 to 53.61 and Event@seg from 39.74 to 47.62 without retraining. Inference takes 3.16 ms per video after feature encoding.
\end{itemize}

\section{False Co-Activation}
\label{sec:fca_analysis}

Let a video be divided into $T$ aligned audio and visual segments, and let $\mathcal{Y}$ be the queried event vocabulary. For modality $m\in\{a,v\}$, $Y_m(t)\subseteq\mathcal{Y}$ is the ground-truth event set in segment $t$. A frozen modality encoder produces $z_m(t)$, and the corresponding frozen text encoder maps the candidate names to
\[
C_m=[c_m^1,\ldots,c_m^{|\mathcal{Y}|}].
\]
The dictionary is encoded once and reused across videos. Dense training-free inference scores each candidate independently as
\[
s_m^{\mathrm{cos}}(t,c)=\cos(z_m(t),c_m^c).
\]
Let $q_m(t,c)$ denote the value that a method assigns to candidate $c$ before making its final prediction. For example, after candidate filtering, $q_m(t,c)$ is the filtered score of each remaining candidate.

\refstepcounter{definition}
\label{def:false_coactivation}
\noindent\textbf{Definition~\thedefinition\ (False Co-Activation).}
Suppose a method retains at least one ground-truth label in a segment. A retained label $j$ that is not in the ground truth is a false co-activation when
\begin{equation}
\label{eq:threshold_inseparable_fca}
j\notin Y_m(t),
\qquad
q_m(t,j)
\geq
\min_{\substack{i\in Y_m(t)\\ i\,\text{retained}}}q_m(t,i).
\end{equation}
Candidate $j$ is at or above the lowest-valued ground-truth label that remains. A scalar cutoff on the same values therefore cannot remove $j$ while keeping every remaining ground-truth label. Definition~\ref{def:false_coactivation} includes the more severe case in which $j$ outranks every remaining ground-truth label.

FCA is a score-level diagnostic rather than a synonym for a final false positive. A class-specific cutoff can keep the candidate out of the final prediction, but the FCA remains in the pre-readout values.

In a joint sparse fit, correlated labels no longer receive shared evidence separately. Once one label explains their common direction, another is selected only if its remaining contribution justifies an additional selection cost. We next analyze the smallest sparse problem that contains this competition, with one annotated event and one competing candidate.

\refstepcounter{lemma}
\label{lem:pairwise_residual}
\noindent\textbf{Lemma~\thelemma\ (Two-Label Sparse Competition).}
Consider a centered unit embedding $z$ (centering is defined in Section~\ref{sec:centering}), a text atom $c_i$ for an annotated event, and a text atom $c_j$ for a candidate outside the annotated set. Their two-label non-negative sparse fit is
\begin{equation}
\label{eq:pairwise_sparse}
w^*
=
\arg\min_{w\geq0}
\left\|z-[c_i,c_j]w\right\|_2^2
+\lambda_0\|w\|_1.
\end{equation}
Let $\rho=c_i^\top c_j$ with $0\leq\rho<1$, and order the atoms so that $c_i^\top z\geq c_j^\top z$. We restrict the analysis to non-negative correlation because the shadow/rival diagnostic concerns labels that share a positively aligned text direction. Negatively correlated atom pairs fall outside this diagnostic. Define $\gamma=\lambda_0/2$ and assume $c_i^\top z>\gamma$. The solution selects both atoms exactly when
\begin{equation}
\label{eq:residual_condition}
\Delta_{j\mid i}(z)
:=(c_j-\rho c_i)^\top z-\gamma(1-\rho)>0.
\end{equation}
If $\Delta_{j\mid i}(z)\leq0$, the fit selects only $c_i$ and removes candidate $c_j$. Appendix~\ref{app:pairwise_proof} gives the proof.

\refstepcounter{definition}
\label{def:fca_types}
\noindent\textbf{Definition~\thedefinition\ (Shadow and Rival Co-Activations).}
Take an FCA $j$. We compare it with each related annotated event $i$ that has at least as much dense evidence and can enter the sparse fit. These reference events form
\[
\mathcal{I}_j(z)
=
\left\{i\in Y_m(t):
c_i^\top z\geq c_j^\top z,
\ c_i^\top z>\gamma,
\ 0<c_i^\top c_j<1
\right\}.
\]
The lemma permits $\rho=0$, but the diagnostic below requires $\rho>0$ because a shadow specifically denotes evidence shared by related atoms. For each $i\in\mathcal{I}_j(z)$, consider the two-label sparse problem in Equation~\eqref{eq:pairwise_sparse}. We define
\[
j\text{ is a shadow}
\quad\Longleftrightarrow\quad
\exists i\in\mathcal{I}_j(z):
\Delta_{j\mid i}(z)\leq0,
\]
and call it a rival otherwise. In Figure~\ref{fig:motivation}(a), \textit{Dog} remains selected while \textit{Cat} disappears, so \textit{Cat} is a shadow. A rival is any FCA for which no eligible stronger annotated event removes the candidate in this two-label analysis. This includes $\mathcal{I}_j(z)=\varnothing$, where no such reference event exists. We report this split only on multi-label segments to study a false candidate competing with several simultaneously annotated events rather than an ordinary single-label ranking error.

Lemma~\ref{lem:pairwise_residual} is exact only for two labels. With a full dictionary, several labels can jointly explain the residual and remove a rival FCA. Shadow and rival are therefore diagnostic tags, not hidden ground-truth causes or a one-to-one assignment to Stages~1 and~2. Appendix~\ref{app:residual_condition} tests how the two-label condition relates to the full 25-label solve.

\section{Method}
\label{sec:method}

\subsection{Problem Formulation}

Given $T$ aligned audio and visual segments and a queried event vocabulary $\mathcal{Y}$, training-free AVEP predicts three binary decisions $\hat y_a(t,c)$, $\hat y_v(t,c)$, and $\hat y_{av}(t,c)$ for every segment $t$ and candidate $c$. The first two indicate event presence in the audio and visual branches, while $\hat y_{av}$ denotes an event that is jointly audible and visible. The method receives the frozen segment embeddings $z_m(t)$ and text dictionaries $C_m$ defined in Section~\ref{sec:fca_analysis}, and uses no task-specific parameter training or segment labels.

\begin{figure}[t]
    \centering
    \includegraphics[width=\linewidth,trim={0pt 5pt 0pt 38pt},clip]{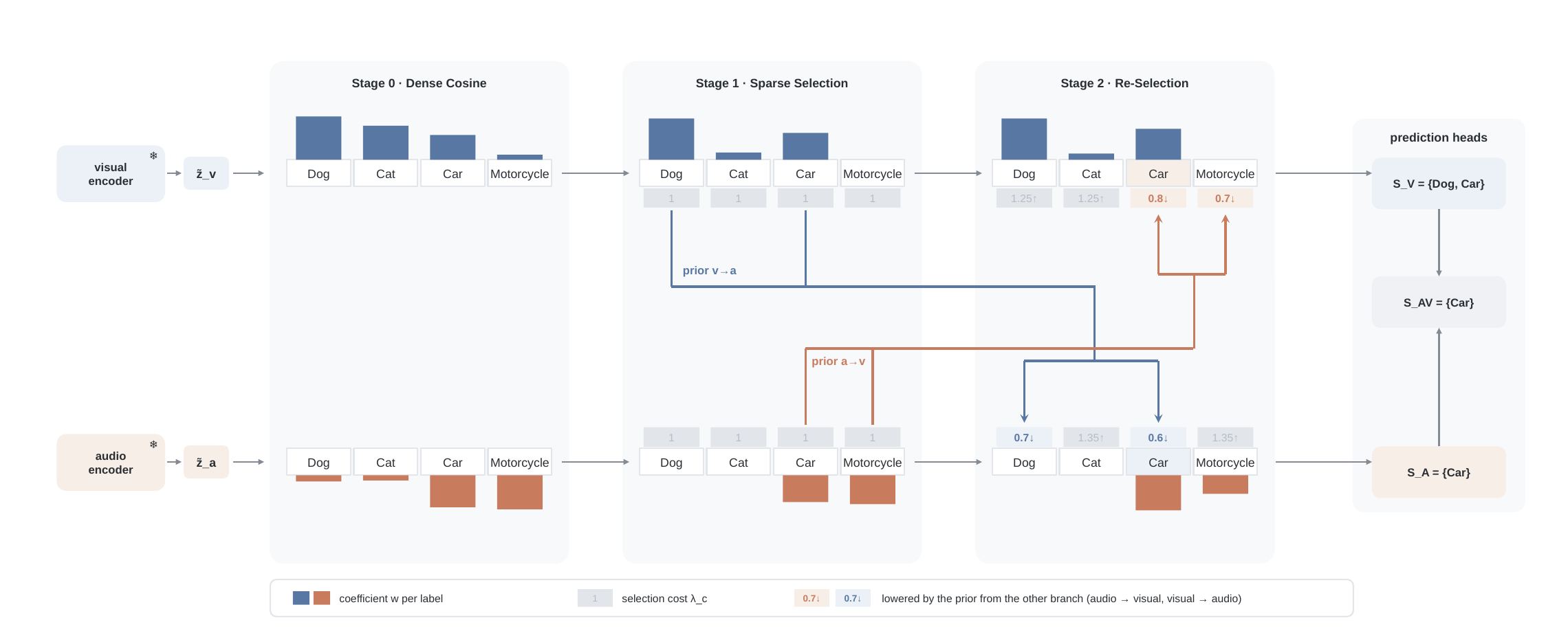}
    \caption{SCoPE on the example from Figure~\ref{fig:motivation}. Stage~1 selects sparse support in each branch. Stage~2 lowers the selection costs of source-supported labels and re-solves from the target embedding. The AV head fuses coefficients retained by both branches.}
    \label{fig:architecture}
\end{figure}

SCoPE has two stages. Stage~1 fits the full event dictionary separately in each modality, making the candidates compete for the segment evidence. Stage~2 uses support from the other modality to change candidate costs and then re-solves from the target embedding. Figure~\ref{fig:architecture} gives the full pipeline.

\subsection{Centered Event Dictionary}
\label{sec:centering}

Mean subtraction~\cite{mu2018all} is a standard first step for removing common components from embedding spaces. Contrastive multimodal encoders~\cite{liang2022mind} also occupy modality-dependent offset regions. We therefore center and renormalize each modality before non-negative sparse inference:
\begin{equation}
\label{eq:centering}
\tilde z_m(t)
=\frac{z_m(t)-\mu_m}{\|z_m(t)-\mu_m\|_2},
\qquad
\tilde c_m^c
=\frac{c_m^c-\bar\mu_m}{\|c_m^c-\bar\mu_m\|_2}.
\end{equation}
Here $\mu_m$ is estimated from an external reference set and $\bar\mu_m$ is the mean of the queried text atoms. We write $\tilde C_m$ for the centered dictionary. Appendix~\ref{app:reproducibility} specifies the references and Appendix~\ref{app:mean_centering_ablation} tests sensitivity to the reference datasets.

\subsection{Stage 1: Competitive Sparse Selection}
\label{sec:stage1}

Stage~1 makes all candidate labels compete to explain the same centered segment embedding. It solves non-negative Lasso independently in each modality:
\begin{equation}
\label{eq:stage1}
w_m(t)
=
\arg\min_{w\geq0}
\left\|\tilde z_m(t)-\tilde C_mw\right\|_2^2
+\lambda_0\|w\|_1.
\end{equation}
Equation~\eqref{eq:stage1} differs from dense matching in how evidence is assigned. Cosine matching compares every text atom with the original embedding. The joint fit instead keeps an additional atom only when it reduces the remaining reconstruction error enough to justify another nonzero coefficient. Correlated labels therefore compete for shared evidence. This is the effect studied by the two-label diagnostic, but the diagnostic does not determine the outcome of the full Stage~1 solve.

The selected support is
\[
S_m(t)=\{c:w_m(t,c)>\varepsilon_{\mathrm{supp}}\}.
\]
Equation~\eqref{eq:stage1} extends the two-atom fit in Equation~\eqref{eq:pairwise_sparse} to all queried labels. Only this sparse, coefficient-weighted support is passed across modalities.

\subsection{Stage 2: Cross-Modal Cost Prior and Target Re-Selection}
\label{sec:prior}
\label{sec:stage2}

Stage~2 uses support from the other modality to reconsider candidates that Stage~1 alone cannot resolve. It exchanges a pooled soft prior, not dense scores or binary predictions. We describe visual-to-audio guidance. The reverse direction uses the same construction with its own strength $\eta_{a\rightarrow v}$.

Figure~\ref{fig:motivation}(b) illustrates the distinction. Visual support can lower the audio selection costs of \textit{Dog} and \textit{Car}, but it does not add either visual score to the audio coefficients. Audio Stage~2 re-solves from its own embedding, so \textit{Car} can gain support without forcing the visually observed but silent \textit{Dog} into the output.

\noindent\textbf{Reconstruction-weighted source prior.}
We measure how well the Stage~1 reconstruction fits each source segment and pool its normalized coefficients:
\begin{equation}
\label{eq:prior_strength}
r_v(t)
=\left[\cos(\tilde z_v(t),\tilde C_vw_v(t))\right]_+,
\qquad
P_v(c)
=\frac{1}{T}\sum_{t=1}^{T}r_v(t)
\frac{w_v(t,c)}{\lVert w_v(t)\rVert_\infty+\varepsilon_{\mathrm{norm}}}.
\end{equation}
Here, $[x]_+=\max(0,x)$ and $\varepsilon_{\mathrm{norm}}$ is a numerical stabilizer. We set $r_v(t)=0$ when $w_v(t)$ has empty support.
The prior retains class identity while allowing evidence at one time to guide another. The factor $r_v(t)$ gives less weight to poorly reconstructed source segments. We divide by the video length rather than by $\sum_t r_v(t)$, so consistently poor reconstruction also reduces the total guidance. If every $r_v(t)$ is zero, Stage~2 uses the Stage~1 cost.
The infinity-norm normalization sets the largest coefficient in each nonempty segment to one, making source strength comparable across time while preserving the relative coefficient sizes within that segment.

\noindent\textbf{Fixed-budget event selection costs.}
The source prior first defines an unnormalized audio cost multiplier
\begin{equation}
\label{eq:weighted_cost}
a_{a,c}=\exp\!\left(-\eta_{v\rightarrow a}P_v(c)\right),
\qquad
\lambda_{a,c}
=\lambda_0
\frac{a_{a,c}}{\frac{1}{|\mathcal{Y}|}\sum_j a_{a,j}}.
\end{equation}
The normalization keeps every cost positive and enforces
$\sum_c\lambda_{a,c}=|\mathcal{Y}|\lambda_0$.
The prior therefore redistributes a fixed budget rather than weakening sparsity as a whole. Stronger source support lowers a class's relative cost, while the other costs rise to preserve the mean. These costs are shared across the target video, but every segment is re-solved from its own embedding.

\noindent\textbf{Target re-selection.}
The audio branch then re-solves its own weighted NN-Lasso problem:
\begin{equation}
\label{eq:stage2}
w_a^{\mathrm{new}}(t)
=\arg\min_{w\geq0}
\left\|\tilde z_a(t)-\tilde C_aw\right\|_2^2
+\sum_c\lambda_{a,c}|w_c|.
\end{equation}
The reverse direction produces $w_v^{\mathrm{new}}(t)$. A lower cost can help the target reconsider a weak label, but it cannot insert the source label or guarantee a correct selection. The unchanged reconstruction term still determines whether the target retains the candidate.

\subsection{Sparse Readout}
\label{sec:calibration}

The sparse coefficients can vary in absolute scale across segments, so a single calibrated cutoff is not appropriate for every output~\cite{lipton2014optimal}. We instead use a classical one-dimensional gap split~\cite{gower1969minimum}, equivalent to cutting the longest edge of single-linkage clustering on the sorted coefficient values. Let the positive Stage~2 coefficients be sorted as
$w_{m,(1)}^{\mathrm{new}}(t)\geq\cdots\geq w_{m,(K)}^{\mathrm{new}}(t)>0$
and append the zero anchor $w_{m,(K+1)}^{\mathrm{new}}(t)=0$. We select the largest gap:
\begin{equation}
\label{eq:active_readout}
k_m^*(t)
=\min\!\left(
\operatorname*{arg\,max}_{1\leq k\leq K}
\left[w_{m,(k)}^{\mathrm{new}}(t)-w_{m,(k+1)}^{\mathrm{new}}(t)\right]
\right),
\qquad
\hat Y_m(t)=\{c_{(1)},\ldots,c_{(k_m^*(t))}\}.
\end{equation}
We set $\hat y_m(t,c)=\mathbb{I}[c\in\hat Y_m(t)]$. The outer minimum retains the smaller support when gaps tie. Empty support produces no prediction, and singleton support retains its label. The zero anchor lets the rule keep the full support when the final drop is strictly largest. The readout has no calibration threshold or support-size parameter.

For audio-visual prediction, define
$S_m^{\mathrm{new}}(t)=\{c:w_m^{\mathrm{new}}(t,c)>\varepsilon_{\mathrm{supp}}\}$.
We retain the support intersection and fuse only its coefficients:
\begin{equation}
\label{eq:av_sparse_readout}
S_{av}(t)=S_a^{\mathrm{new}}(t)\cap S_v^{\mathrm{new}}(t),
\qquad
w_{av}(t,c)
=\mathbb{I}[c\in S_{av}(t)]
\left[\alpha w_a^{\mathrm{new}}(t,c)+(1-\alpha)w_v^{\mathrm{new}}(t,c)\right].
\end{equation}
Applying the same sparse readout to $w_{av}$ gives $\hat y_{av}$. For comparability, we inherit AV$^2$A's CLIP+CLAP fusion weight~\cite{shaar2025av2a} and do not retune it by dataset. The audio and visual outputs remain unchanged by this AV fusion. Appendix~\ref{app:reproducibility} gives solver and temporal decoding details.

\section{Experiments}
\label{sec:experiments}

\subsection{Setup}
\label{sec:setup}

\noindent\textbf{Datasets and metrics.}
LLP~\cite{tian2020avvp} contains 25 event classes and modality-wise segment annotations for audio, visual, and audio-visual events. We report the standard segment-level and event-level F1 scores for Audio, Visual, Audio-Visual, Type, and Event. OV-AVEBench~\cite{zhou2025ovavel} contains 67 queried classes, for which we report chunk accuracy, segment F1, event F1, and their average. VGGSound-AVEL100k uses 141 VGGSound~\cite{chen2020vggsound} event classes and evaluates temporal event localization using segment micro F1 and event F1. Appendix~\ref{app:dataset_protocols} gives the split, temporal annotation, evaluated sample, and metric details.

\noindent\textbf{Frozen encoders and dictionaries.}
The primary experiments use CLIP ViT-B/32 for vision and LAION-CLAP HTSAT-tiny for audio. For CLIP+CLAP, we follow AV$^2$A's released implementation. CLAP uses \texttt{This is a sound of \{label\}}, and CLIP uses \texttt{A \{label\}}. LanguageBind uses \texttt{A \{label\}} in both modalities. ImageBind uses the same modality-specific templates as CLIP+CLAP. LLP uses 25 atoms, OV-AVEBench evaluates 67 event classes, and VGGSound-AVEL100k uses 141 atoms. Each dictionary is encoded once and reused for every video. Within each table block, compared methods use the same frozen encoders, prompt templates, and vocabulary. We evaluate SCoPE with both CLIP+CLAP and LanguageBind on LLP, and with ImageBind on OV-AVEBench and VGGSound-AVEL100k. The CLIP+CLAP OV-AVEBench block uses exactly the 67 evaluated event prompts. The ImageBind base and SCoPE rows share those prompts plus an auxiliary \texttt{other} prompt, which is treated as background during evaluation.

\noindent\textbf{References and implementation.}
We compute external feature means from MS-COCO 2017 for vision and ESC-50 for audio. Section~\ref{sec:method} specifies the inference settings. The base rows in Table~\ref{tab:avel_transfer_results} use the same frozen dense decoder across datasets. The ImageBind agreement row applies OV-AVEL's A/V agreement rule to the same dense scores. We quote the published LLP Dense and AV$^2$A rows and reproduce the OV-AVEBench AV$^2$A row from its released predictions with the official evaluator. Appendix~\ref{app:reproducibility} gives the evaluation populations, solver, and preprocessing details.

\noindent\textbf{Hyperparameters.}
On the 563 official LLP validation videos with complete frozen feature caches, we select the sparse cost and directional prior strengths. This gives $\lambda_0=0.30$, $\eta_{v\rightarrow a}=16$, and $\eta_{a\rightarrow v}=4$. We use support tolerance $\varepsilon_{\mathrm{supp}}=10^{-6}$ and normalization stabilizer $\varepsilon_{\mathrm{norm}}=10^{-8}$. The AV fusion weight $\alpha=0.45$ is inherited from AV$^2$A rather than selected on this subset. We freeze these values for LLP test evaluation and transfer them unchanged to the other backbones and datasets. Appendix~\ref{app:dataset_protocols} reports the search grid, while Appendix~\ref{app:hyperparameter_sensitivity} reports sensitivity to the fixed settings.

\noindent\textbf{Runtime protocol.}
We report cached-feature inference separately from the 393 ms encoding reference used by OV-AVEL. The AV$^2$A latency is the single-GPU equivalent of its reported 4$\times$TITAN RTX run. Appendix~\ref{app:reproducibility} gives the measured scope and hardware.

\subsection{Main Results}
\label{sec:main_results}

% Final-method release: paper_values_release.json (2026-08-07).
\begin{table}[t]
\centering
\caption{Training-free audio-visual video parsing on LLP (F1, \%). LanguageBind follows AV$^2$A's additional-backbone evaluation.}
\label{tab:main_results}
\vspace{3pt}
\resizebox{\linewidth}{!}{
\begin{tabular}{lcccccccccc}
\toprule
\multirow{2}{*}{Method}
& \multicolumn{2}{c}{Audio}
& \multicolumn{2}{c}{Visual}
& \multicolumn{2}{c}{Audio-Visual}
& \multicolumn{2}{c}{Type}
& \multicolumn{2}{c}{Event} \\
\cmidrule(lr){2-3}
\cmidrule(lr){4-5}
\cmidrule(lr){6-7}
\cmidrule(lr){8-9}
\cmidrule(lr){10-11}
& Seg & Evt & Seg & Evt & Seg & Evt & Seg & Evt & Seg & Evt \\
\midrule
CLIP+CLAP~\cite{radford2021clip,elizalde2023clap}
& 17.6 & 15.1 & 18.8 & 17.8 & 30.0 & 25.6 & 22.1 & 19.5 & 18.3 & 16.2 \\
\quad with AV$^2$A
& 32.7 & 28.5 & 43.5 & 41.5 & \textbf{55.1} & \textbf{48.2} & 43.8 & 39.4 & 34.7 & \textbf{31.2} \\
\quad with SCoPE (Stage 1)
& 32.42 & 23.98 & 50.91 & 42.29 & 42.67 & 36.41 & 42.00 & 34.23 & 31.61 & 23.17 \\
\quad with \textbf{SCoPE (Stage 2)}
& \textbf{40.32} & \textbf{31.47}
& \textbf{59.06} & \textbf{52.39}
& 54.37 & 48.05
& \textbf{51.25} & \textbf{43.97}
& \textbf{39.74} & 31.04 \\
\midrule
LanguageBind~\cite{zhu2024languagebind}
& 20.3 & 17.5
& 21.3 & 20.0
& 24.3 & 21.0
& 22.0 & 19.5
& 20.8 & 18.4 \\
\quad with AV$^2$A
& 40.9 & 35.9
& 57.4 & 54.4
& \textbf{59.1} & \textbf{52.3}
& 52.4 & 47.5
& 43.4 & \textbf{38.9} \\
\quad with SCoPE (Stage 1)
& 31.45 & 18.90
& 62.38 & 55.87
& 49.08 & 41.84
& 47.64 & 38.87
& 36.48 & 24.98 \\
\quad with \textbf{SCoPE (Stage 2)}
& \textbf{45.81} & \textbf{37.21}
& \textbf{64.61} & \textbf{59.47}
& 58.52 & 52.31
& \textbf{56.31} & \textbf{49.66}
& \textbf{46.01} & 37.35 \\
\bottomrule
\end{tabular}
}
\end{table}

% Final-method release: paper_values_release.json (2026-08-07).
\begin{table}[t]
\centering
\caption{Training-free event localization across datasets and frozen backbones (\%).}
\label{tab:avel_transfer_results}
\small
\setlength{\tabcolsep}{2.9pt}
\begin{tabular}{@{}lccccccc@{}}
\toprule
\multicolumn{1}{c}{} & \multicolumn{5}{c}{OV-AVEBench} & \multicolumn{2}{c}{VGGSound-AVEL100k} \\
\cmidrule(lr){2-6}\cmidrule(lr){7-8}
Method & Acc. & Seg-F1 & Event-F1 & Avg. & Unseen Avg. & Seg F1 & Event F1 \\
\midrule
CLIP+CLAP~\cite{radford2021clip,elizalde2023clap} & 44.46 & 37.57 & 30.89 & 37.64 & 38.17 & 19.09 & 6.09 \\
\quad with OV-AVEL~\cite{zhou2025ovavel} & 51.86 & 42.92 & 32.85 & 42.54 & 42.05 & 26.98 & 14.83 \\
\quad with AV$^2$A & 43.18 & 31.90 & 24.34 & 33.14 & 32.69 & 28.47 & 24.21 \\
\quad with \textbf{SCoPE} & \textbf{57.11} & \textbf{49.81} & \textbf{45.06} & \textbf{50.66} & \textbf{51.21} & \textbf{49.63} & \textbf{30.88} \\
\midrule
ImageBind~\cite{girdhar2023imagebind} & 36.98 & 30.61 & 24.53 & 30.71 & 30.53 & 15.46 & 5.43 \\
\quad with OV-AVEL~\cite{zhou2025ovavel} & 59.41 & 48.07 & 35.66 & 47.71 & 48.14 & 45.02 & 22.97 \\
\quad with \textbf{SCoPE} & \textbf{60.58} & \textbf{53.01} & \textbf{48.11} & \textbf{53.90} & \textbf{55.69} & \textbf{56.92} & \textbf{38.69} \\
\bottomrule
\end{tabular}
\end{table}

\noindent\textbf{LLP.}
Table~\ref{tab:main_results} compares training-free methods using the same frozen backbones and vocabularies within each block. With CLIP+CLAP, Stage~2 raises Audio, Visual, and Audio-Visual segment F1 over Stage~1 by 7.90, 8.15, and 11.70 points. Compared with the reported AV$^2$A values, it raises Type@seg from 43.8 to 51.25 and Event@seg from 34.7 to 39.74. With LanguageBind, the same fixed configuration reaches 56.31 Type@seg and 46.01 Event@seg.

\noindent\textbf{Benchmark and backbone transfer.}
Table~\ref{tab:avel_transfer_results} evaluates the same frozen SCoPE configuration on larger vocabularies and alternative backbones. On OV-AVEBench, SCoPE reaches 50.66 Avg. with CLIP+CLAP and 53.90 with ImageBind, including 51.21 and 55.69 on unseen classes. On strict VGGSound-AVEL100k, SCoPE reaches 49.63 segment F1 and 30.88 event F1 with CLIP+CLAP, compared with 28.47 and 24.21 for the AV$^2$A baseline. With ImageBind, OV-AVEL-style agreement raises the dense baseline from 15.46 to 45.02 segment F1, and SCoPE further reaches 56.92. Event F1 rises from 5.43 to 22.97 and 38.69, respectively.

\subsection{Mechanism Analysis}
\label{sec:mechanism_analysis}

Tables~\ref{tab:hard_exchange} and~\ref{tab:runtime_main} summarize the controlled exchange variants and inference cost, respectively.

\begin{center}
\begin{minipage}{0.84\linewidth}
\vspace{0pt}
\centering
\centering
\resizebox{\linewidth}{!}{%
\begin{tabular}{@{}lccccc@{}}
\toprule
Variant & \shortstack{Soft\\prior} & \shortstack{Reconstruction\\weighting} & \shortstack{Target\\re-selection} & Type@seg & Event@seg \\
\midrule
No exchange & $\times$ & $\times$ & $\times$ & 42.00 & 31.61 \\
Hard prior & $\times$ & \checkmark & \checkmark & 50.35 & 39.21 \\
Direct fusion & \checkmark & \checkmark & $\times$ & 44.64 & 32.04 \\
Soft prior, unweighted & \checkmark & $\times$ & \checkmark & 50.99 & 39.68 \\
SCoPE & \checkmark & \checkmark & \checkmark & \textbf{51.25} & \textbf{39.74} \\
\bottomrule
\end{tabular}%
}

\par\smallskip
\refstepcounter{table}
\label{tab:hard_exchange}
\parbox{\linewidth}{\raggedright\footnotesize\textbf{Table~\thetable:} Cross-modal prior ablation on LLP (F1, \%).}
\end{minipage}

\medskip

\begin{minipage}{0.96\linewidth}
\vspace{0pt}
\centering
\centering
\footnotesize
\begin{minipage}[t]{0.56\linewidth}
\vspace{0pt}
\centering
\setlength{\tabcolsep}{4.2pt}
\begin{tabular}[t]{@{}lrrr@{}}
\toprule
\multicolumn{4}{c}{OV-AVEBench runtime (ms/video)} \\
\cmidrule(lr){1-4}
Method & Encoding & Inference & Total \\
\midrule
OV-AVEL & 393 & 4.80 & 397.80 \\
AV$^2$A & 393 & 1174 & 1567 \\
SCoPE & 393 & \textbf{3.16} & \textbf{396.16} \\
\bottomrule
\end{tabular}
\end{minipage}\hfill
\begin{minipage}[t]{0.40\linewidth}
\vspace{0pt}
\centering
\setlength{\tabcolsep}{4.2pt}
\begin{tabular}[t]{@{}lrrrr@{}}
\toprule
\multicolumn{5}{c}{NN-Lasso scaling (ms/segment)} \\
\cmidrule(lr){1-5}
Labels $K$ & 61 & 1k & 5k & 10k \\
Latency & 0.17 & 0.18 & 0.34 & 0.65 \\
\bottomrule
\end{tabular}
\end{minipage}

\par\smallskip
\refstepcounter{table}
\label{tab:runtime_main}
\parbox{\linewidth}{\raggedright\footnotesize\textbf{Table~\thetable:} Runtime and vocabulary scaling.}
\end{minipage}
\end{center}

\noindent\textbf{What should be exchanged, and how?}
All rows start from identical Stage~1 coefficients and vary three choices: prior form, reconstruction weighting, and target re-selection. A hard prior gives every selected source label the same value, whereas a soft prior gives stronger selected labels more influence. Direct fusion adds the soft, reconstruction-weighted prior to the target coefficients without a new solve. Its directional ratio is fixed to SCoPE's 4:1 setting, and only one global scale is selected on validation. The unweighted soft-prior row keeps target re-selection but lets selected segments contribute equally. SCoPE uses the soft prior, reconstruction weighting, and target re-selection together.

For the unweighted row, we rescale $\eta$ using validation features without labels so its mean absolute selection-cost change matches SCoPE. This isolates reconstruction weighting from overall prior strength.

SCoPE gives the strongest result on both metrics. Target re-selection provides the clearest controlled gain. Direct fusion trails SCoPE by 6.61 Type@seg points and 7.70 Event@seg points. The soft prior adds 0.90 Type@seg points over the hard prior, and reconstruction weighting adds 0.26 over the matched unweighted prior. Appendix~\ref{app:qualitative_examples} provides clip-level examples of target re-selection.

\begin{wrapfigure}{r}{0.39\textwidth}
    \vspace{-0.8\baselineskip}
    \centering
    \includegraphics[width=\linewidth]{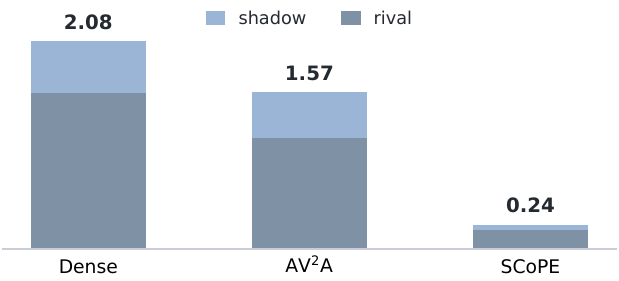}
    \par\smallskip
    \refstepcounter{figure}
    \label{fig:av_fca}
    \parbox{\linewidth}{\raggedright\small Figure~\thefigure: Mean AV false co-activations per multi-label LLP segment. Each label inherits its branch-level shadow or rival tag.}
    \vspace{-0.8\baselineskip}
\end{wrapfigure}
\noindent\textbf{Boundary of the Audio-Visual readout.}
AV$^2$A's score fusion can activate an Audio-Visual event from one strong branch, whereas SCoPE requires support in both Stage~2 branches. Replacing AV$^2$A's fusion with the strict intersection of its branch predictions lowers Audio-Visual segment F1 from 55.1 to 51.8. SCoPE reaches 54.37 while retaining its both-branch support rule, recovering 2.57 points over strict intersection. Among the predictions added by AV$^2$A's fusion beyond strict intersection, 58.5\% are incorrect.

Figure~\ref{fig:av_fca} aggregates the same boundary over multi-label AV segments.\footnote{Dense methods do not define a separate AV score, so we fuse their branch cosine scores with the AV readout weight $\alpha$. An AV FCA is tagged as a shadow if either modality branch gives it that tag, and otherwise as a rival.} Score fusion retains more one-sided labels, whereas SCoPE requires support from both modalities.

\subsection{Efficiency}
\label{sec:efficiency}

SCoPE solves NN-Lasso only at inference time over a text dictionary encoded once. Over all 5{,}820 OV-AVEBench videos, the cached-feature pipeline takes 3.16 ms per video, or 316 videos/s. Table~\ref{tab:runtime_main} adds this measurement to the shared encoding reference. The 396.16 ms total is therefore an arithmetic composite rather than a separately timed end-to-end run. We also measure visual NN-Lasso latency at 61, 1k, 5k, and 10k labels using cached LLP features and 200 FISTA iterations. With batches of 128 segments, latency stays below 0.35 ms per segment through 5k labels and reaches 0.65 ms with 10k labels. At batch size 1, the 10k-label solve takes 0.022 s per segment.

\section{Related Work}

\subsection{Fixed-Vocabulary AVEP}

Audio-visual event localization~\cite{tian2018avel} identifies events that are both audible and visible in temporal video segments. Audio-visual video parsing~\cite{tian2020avvp} broadens this task to audio-only, visual-only, and jointly audio-visual events under video-level supervision. Later methods improve temporal localization and modality assignment with task-specific models~\cite{wu2021heterogeneous,fan2023language,lai2023valor,yu2025prefm,gao2023cmpae,yang2026hschg}. Because these models are trained for a fixed label set, extending them to new event names or domains requires new annotations or optimization.

\subsection{Open-Vocabulary AVEP}

Open-vocabulary AVEP queries events through text rather than a fixed classifier. OV-AVEL~\cite{zhou2025ovavel} introduced OV-AVEBench and a CLIP+CLAP baseline that applies fixed thresholds to frozen text-matching scores. AV$^2$A~\cite{shaar2025av2a} adds video-level candidate filtering and adaptive class thresholds, and subsequent training-free work~\cite{cho2026cue2rule} further adapts score thresholds. These methods refine decisions over frozen text-matching scores. SCoPE intervenes at a different point by first forming sparse, competitive supports and then exchanging them across modalities.

\subsection{Sparse Event Inference}

Sparse coding~\cite{lee2006efficient,mairal2010online} explains an observation with a compact set of dictionary atoms that compete for shared evidence. Recent work~\cite{bhalla2024splice,zhang2025transformation} applies this idea to foundation-model embeddings by decomposing CLIP features over text-defined concepts and constructing sparse concept representations for audio features. These methods analyze each modality independently. SCoPE instead indexes the audio and visual dictionaries with the same event vocabulary. This shared label space lets source support redistribute a fixed selection-cost budget in the target, which then solves its own sparse problem again. Sparse selection therefore provides both within-modality competition and cross-modal event guidance.

\section{Discussion and Conclusion}
\label{sec:discussion}

\noindent\textbf{Future directions.}
SCoPE uses a video-level prior and offline temporal post-processing. Extending both to causal inference is a natural next step. Adaptive costs and stronger branch evidence from source separation or object localization are also promising directions that preserve training-free inference.

\noindent\textbf{Limitations.}
Lemma~\ref{lem:pairwise_residual} is exact only for two atoms and does not completely characterize full-dictionary selection. Stage~2 requires useful source support and target evidence, but does not know whether the source is correct. Encoder confusion or missing target evidence can therefore leave false activations or missed events. The sparse readout can remove weak events when event strengths differ substantially. Directional prior strengths and the conservative AV intersection are fixed operating points that may vary by deployment.

\noindent\textbf{Conclusion.}
SCoPE replaces independent label scoring with sparse competition and target-controlled cross-modal re-selection. Stage~1 suppresses related candidates by making them compete for the same evidence. Stage~2 pools reconstruction-weighted source support, but the target reconstruction still decides whether a candidate is selected. The AV head makes a separate precision--recall trade-off. Support intersection suppresses false AV segment--class entries but can miss events with one-sided evidence.

\clearpage
\bibliographystyle{iclr2027_conference}
\bibliography{references}

\clearpage
\appendix
\raggedbottom

\section*{Supplementary Material}

\section{Sparse Candidate and Event Analysis}
\label{app:dictionary}
\subsection{Pairwise Residual Criterion}
\label{app:residual_condition}

Lemma~\ref{lem:pairwise_residual} asks whether a competing candidate retains evidence after the direction explained by a stronger annotated event is removed. We first prove the result, then specify its use as a fixed diagnostic with the full dictionary.

\phantomsection
\label{app:pairwise_proof}
\noindent\textit{Proof.}
Suppose $c_j$ is inactive. The one-atom optimum for $c_i$ under Equation~\eqref{eq:pairwise_sparse} is
\[
w_i^\star=c_i^\top z-\gamma,
\qquad
\gamma=\lambda_0/2.
\]
The KKT condition keeps $c_j$ inactive exactly when
\[
c_j^\top(z-w_i^\star c_i)\leq\gamma.
\]
Substituting $w_i^\star$ and using $\rho=c_i^\top c_j$ gives
\[
(c_j-\rho c_i)^\top z\leq\gamma(1-\rho).
\]
When this inequality is violated, the coefficient of $c_j$ is positive. The score ordering keeps the coefficient of $c_i$ at least as large, so the pairwise solution selects both atoms. Otherwise, it selects only $c_i$. Since $\rho<1$, the two-atom objective is strictly convex on their span and the solution is unique.
\hfill$\square$

For the idealized mixture $z=ac_i+bc_j$ with $a,b\geq0$, dense matching assigns $c_j$ the score $b+\rho a$. After conditioning on $c_i$, the shared term $\rho a$ cancels, and Equation~\eqref{eq:residual_condition} reduces to
\[
b>\frac{\gamma}{1+\rho}.
\]
The second atom therefore survives only when its independent mixture weight exceeds the event selection cost implied by its similarity to the selected atom.

\paragraph{Pairwise diagnostic tags.}
For every candidate $j$ outside the annotated event set in a multi-label segment, we compute one tag from the centered target embedding and text dictionary. Let $s_k=c_k^\top z$ and $\rho_{ij}=c_i^\top c_j$. We collect all annotated reference events $i$ with $s_i\geq s_j$, $s_i>\gamma$, and $\rho_{ij}>10^{-8}$. If no such reference exists, $j$ is a rival. Otherwise, we evaluate
\[
\delta_{ij}=s_j-\rho_{ij}s_i-\gamma(1-\rho_{ij})
\]
for every eligible reference. Candidate $j$ is a shadow when $\min_i\delta_{ij}\leq0$ and a rival otherwise. We record the minimizing reference and margin for audit, but evaluate all eligible references when assigning the tag.

The tag is fixed from the target embedding before methods are compared. Dense matching compares all candidates using centered cosine values, AV$^2$A compares its retained candidates using released branch similarities, and SCoPE compares positive sparse coefficients. Because the methods retain different numbers of ground-truth labels, each score floor is the lowest-valued ground-truth label retained by that method.

This is a pairwise diagnostic rather than an explanation of the full-dictionary prediction. It tests whether any one annotated event can explain the competing candidate under the two-atom model, while the full solve may also depend on interactions with the remaining atoms.

\paragraph{Full-dictionary check.}
The lemma is exact for a two-atom dictionary: the pairwise solution retains both atoms when Equation~\eqref{eq:residual_condition} holds and only the stronger atom otherwise. We compare this exact implication with Stage~1 solutions over all 25 LLP atoms. On segments with exactly two annotated events, we order their atoms by dense score and apply the condition to the lower-scoring atom. This checks whether annotated events remain in the support and is not an accuracy claim for the candidate tags.

\begin{center}
\refstepcounter{table}
\label{tab:full_dictionary_residual}
\begin{minipage}{\linewidth}
\small
\textbf{Table~\thetable:} Pairwise-condition agreement under two-label and full 25-label sparse solves on LLP. Agreement means that the condition correctly predicts whether both annotated events remain selected.
\par\smallskip
\centering
\setlength{\tabcolsep}{4.0pt}
\begin{tabular}{lrr}
\toprule
Dictionary & Audio agreement & Visual agreement \\
\midrule
2 labels & 100.00\% & 100.00\% \\
25 labels & 96.15\% & 99.45\% \\
\bottomrule
\end{tabular}
\end{minipage}
\end{center}

The two-label solve follows Lemma~\ref{lem:pairwise_residual} exactly. With all 25 labels competing, its outcome agrees with the pairwise condition in 96.15\% of audio cases and 99.45\% of visual cases. In the removal direction, where $\Delta_{j\mid i}(z)\leq0$, the lower-scoring event incorrectly remains selected in only 0.55\% of audio cases and no visual cases. Most remaining disagreement occurs in the opposite direction, when additional atoms remove an event that the two-label fit would retain.

\section{Implementation Details}
\label{app:implementation}
\subsection{Reproducibility and Solver Configuration}
\label{app:reproducibility}

SCoPE has no trainable parameters. Equations~\eqref{eq:stage1} and~\eqref{eq:stage2} are solved by non-negative FISTA for 200 fixed iterations from zero initialization. For each centered dictionary, its Lipschitz constant is computed once and reused. Equation~\eqref{eq:prior_strength} uses $\varepsilon_{\mathrm{norm}}=10^{-8}$, and coefficients above $\varepsilon_{\mathrm{supp}}=10^{-6}$ define the numerical support. Stage~2 also starts from zero and differs from Stage~1 only through the class-specific costs in Equation~\eqref{eq:weighted_cost}.

\paragraph{Sparse readout.}
Coefficients above $\varepsilon_{\mathrm{supp}}$ form the numerical support. We sort this support by coefficient value, append a zero anchor, and apply Equation~\eqref{eq:active_readout}. Ties between maximum gaps retain the smaller prefix. Empty supports remain empty, while singleton supports retain their only label. The AV output first intersects the two Stage~2 supports, fuses their coefficients with audio weight $\alpha=0.45$, and applies the same rule.

Each event dictionary is encoded once. For controlled comparisons, every inference variant reuses the same cached segment features. The visual and audio means use 118{,}287 MS-COCO and 1{,}600 ESC-50 embeddings, respectively, while text atoms use the queried-dictionary mean. We use $\lambda_0=0.30$, $\eta_{v\rightarrow a}=16$, and $\eta_{a\rightarrow v}=4$. For LLP, we first apply no gap closing to the audio and visual outputs and one-segment gap closing to the joint AV output. We then remove any class predicted in fewer than two segments over the video. The code includes preprocessing, pooled-prior construction, fixed-budget sparse inference, sparse readout, temporal post-processing, and evaluation.

For runtime, we process the complete 5{,}820-video OV-AVEBench split three times on one NVIDIA TITAN RTX and report the median. The 3.160 ms/video measurement starts with cached features already loaded in memory and includes normalization, centering, both sparse stages, pooled-prior construction, and final readout. It excludes backbone and text encoding, disk loading, evaluation, and artifact writing. The total in Table~\ref{tab:runtime_main} is the arithmetic sum of this measurement and the existing 393 ms encoding reference, not an independently timed end-to-end run. The vocabulary-scaling block uses cached LLP visual segment features on the same GPU and reports synchronized median latency for 200 zero-initialized FISTA iterations at batch size 128. Text-dictionary encoding is excluded.

\subsection{Evaluation Datasets and Protocols}
\label{app:dataset_protocols}

\paragraph{LLP.}
We evaluate both CLIP+CLAP and LanguageBind on all 1{,}109 raw-available LLP test videos. The published Dense and AV$^2$A values in Table~\ref{tab:main_results} were reported on the 1{,}124 test videos available to AV$^2$A. The comparison therefore shares frozen encoders and vocabulary but is not paired by video. Each ten-second video has ten aligned one-second segments and separate 25-class audio and visual annotations. Joint audio-visual labels are derived from their intersection. We use the official AVVP evaluator and report video-averaged segment and event F1 for Audio, Visual, Audio-Visual, Type, and Event.

We select hyperparameters on the 563 LLP validation videos with complete audio and visual features. We evaluate $\lambda_0\in\{0.2,0.3,0.4\}$, geometric prior scale $g\in\{2\sqrt{2},4,4\sqrt{2},8,8\sqrt{2}\}$, and direction ratio $\eta_{v\rightarrow a}:\eta_{a\rightarrow v}\in\{1{:}1,4{:}1\}$. The selected values are $\lambda_0=0.30$, $g=8$, $\eta_{v\rightarrow a}=16$, and $\eta_{a\rightarrow v}=4$. The sparse readout is fixed by design, and the complete configuration is transferred without retuning.

\paragraph{OV-AVEBench.}
We use all 5{,}820 test videos and 67 classes, including 46 seen and 21 unseen classes. Following OV-AVEL, we reduce a multi-label output to the class with the longest predicted duration and report the mean of chunk accuracy, segment F1, and event F1 as Avg. CLIP+CLAP uses the 67 evaluation prompts. ImageBind adds one fixed \textit{other} prompt, which is collapsed to background, for both its base and SCoPE rows.

\paragraph{VGGSound-AVEL100k.}
We use the 19{,}378 of 20{,}216 test videos with complete inputs and feature caches. Every method uses this subset and the same 141-class dictionary. We report segment micro F1 and event F1 averaged over temporal IoU thresholds $\{0.1,0.3,0.5,0.7,0.9\}$.

\paragraph{Assets and licenses.}
We use public datasets and frozen backbones only for research evaluation and feature extraction and do not redistribute raw media or checkpoints. CLIP is released under MIT and LAION-CLAP under CC0-1.0. LLP, OV-AVEBench, VGGSound, MS-COCO, ESC-50, UrbanSound8K, FSD50K, ImageNet, and CIFAR-100 remain subject to their source licenses. The still frame in Figure~\ref{fig:motivation} comes from the public LLP benchmark and is used only to illustrate the benchmark example under its original access conditions.

\section{Additional Results and Diagnostics}
\label{app:ablations}
\subsection{Hyperparameter Sensitivity}
\label{app:hyperparameter_sensitivity}

Figure~\ref{fig:hyperparameter_sensitivity} varies one setting at a time on the LLP validation subset while fixing the others. The reported setting is $\lambda_0=0.3$, geometric prior scale $g=\sqrt{\eta_{v\rightarrow a}\eta_{a\rightarrow v}}=8$, and direction ratio $\eta_{v\rightarrow a}:\eta_{a\rightarrow v}=4{:}1$. The AV fusion weight $\alpha=0.45$ follows AV$^2$A, so its panel is diagnostic rather than a selection curve.

\begin{center}
\begin{minipage}{0.96\linewidth}
    \centering
    \includegraphics[width=\linewidth]{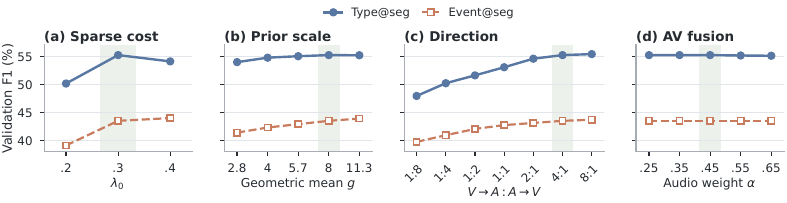}
    \par\smallskip
    \refstepcounter{figure}
    \label{fig:hyperparameter_sensitivity}
    \parbox{\linewidth}{\small Figure~\thefigure: Hyperparameter sensitivity on LLP validation. Curves report Type@seg and Event@seg while varying (a) $\lambda_0$, (b) the geometric mean $g$ of the prior strengths, (c) their direction ratio, and (d) the AV fusion weight $\alpha$. Shading marks the reported setting.}
\end{minipage}
\end{center}

\subsection{Mean Centering Ablation}
\label{app:mean_centering_ablation}

We vary the external datasets used to estimate the modality means while keeping the frozen encoders, candidate dictionary, regularization, cross-modal priors, and readout fixed. Each row changes one reference modality relative to the main ESC-50 and MS-COCO setting. No LLP feature is used to estimate these means.

\begin{table}[h!]
\centering
\caption{Mean-centering reference sensitivity on LLP (\%).}
\label{tab:mean_centering_ablation}
\resizebox{\linewidth}{!}{%
\begin{tabular}{llccccc}
\toprule
Audio mean & Visual mean & A@seg & V@seg & AV@seg & Type@seg & Event@seg \\
\midrule
ESC-50 & MS-COCO & \textbf{40.32} & 59.06 & 54.37 & \textbf{51.25} & \textbf{39.74} \\
UrbanSound8K & MS-COCO & 38.72 & 58.77 & 52.35 & 49.95 & 38.69 \\
FSD50K & MS-COCO & 38.53 & \textbf{60.30} & \textbf{54.45} & 51.09 & 38.27 \\
ESC-50 & ImageNet-1k & 40.22 & 57.98 & 53.23 & 50.47 & 39.55 \\
ESC-50 & CIFAR-100 & 39.49 & 56.14 & 52.34 & 49.32 & 38.52 \\
\bottomrule
\end{tabular}}
\end{table}

This comparison tests whether the result depends on a particular external reference set rather than a mean estimated from LLP or its test distribution.

\subsection{Compatibility with Training-Free Decision Components}
\label{app:component_compatibility}

As a fixed compatibility diagnostic, we apply AV$^2$A-style dynamic thresholds and candidate filtering to the audio, visual, and Audio-Visual outputs of the SCoPE pipeline. Table~\ref{tab:component_compatibility} reports each component separately and together. Neither component requires parameter training, and the reported SCoPE configuration itself remains unchanged.

\begin{table}[h]
\centering
\small
\setlength{\tabcolsep}{7pt}
\caption{Compatibility with dynamic thresholds and candidate filtering on LLP (F1, \%). Components are applied to all three output branches without retraining.}
\label{tab:component_compatibility}
\begin{tabular}{lcc}
\toprule
Variant & Type@seg & Event@seg \\
\midrule
SCoPE & 51.25 & 39.74 \\
$+$ Dynamic thresholds & 52.27 & 42.30 \\
$+$ Filtering & 53.48 & 47.15 \\
$+$ Dynamic thresholds $+$ filtering & \textbf{53.61} & \textbf{47.62} \\
\bottomrule
\end{tabular}
\end{table}

Both components improve SCoPE independently, and their combination reaches 53.61 Type@seg and 47.62 Event@seg. This result shows that sparse prior exchange can be combined with score-level decision rules that act later in the pipeline.

\subsection{Variation by Modality Structure and Event Duration}
\label{app:event_variation}

One fixed configuration does not help every event equally. We first group events by ground-truth modality support. SYM events are annotated in both audio and visual streams, whereas ASYM events are annotated in only one. We then divide ground-truth event runs into short ($\leq5$ seconds) and long ($>5$ seconds) groups. To isolate temporal boundary quality from event detection, the duration analysis compares the best temporal IoU on the same runs for which both stages produce an overlapping prediction.

\begin{table}[h!]
\centering
\caption{Stage~1 to Stage~2 variation on LLP. (a) Event@seg by modality support. (b) Mean temporal IoU on matched event runs by duration. All values are percentages.}
\label{tab:event_variation}
\small
\textbf{(a) Modality support}\par\smallskip
\begin{tabular}{lrrr}
\toprule
Event structure & Stage~1 & Stage~2 & Change \\
\midrule
SYM & 33.1 & \textbf{46.4} & $+13.3$ \\
ASYM & 32.3 & \textbf{35.9} & $+3.6$ \\
\bottomrule
\end{tabular}

\medskip
\textbf{(b) Event duration}\par\smallskip
\begin{tabular}{lrrr}
\toprule
Duration & Stage~1 & Stage~2 & Change \\
\midrule
Short ($\leq5$ s) & \textbf{63.4} & 63.0 & $-0.4$ \\
Long ($>5$ s) & 43.4 & \textbf{64.1} & $+20.7$ \\
\bottomrule
\end{tabular}
\end{table}

Stage~2 raises Event@seg by 13.3 points on SYM events but by only 3.6 points on ASYM events. Mean temporal IoU is nearly unchanged for short runs, whereas it rises from 43.4 to 64.1 for long runs. Cross-modal exchange is therefore most useful when evidence is available in both modalities and persists over time. This variation motivates event-aware costs and temporal decoding as future work.

\subsection{Qualitative Examples}
\label{app:qualitative_examples}

Figures~\ref{fig:qualitative_target_reselection} and~\ref{fig:qualitative_more_examples} visualize six LLP test examples. Figure~\ref{fig:qualitative_target_reselection} first follows two clips in detail. Black cells are ground truth, orange cells are AV$^2$A predictions, blue cells mark positive Stage~1 support, and green cells are the final Stage~2 predictions after target re-selection and sparse readout. Figure~\ref{fig:qualitative_more_examples} then provides four compact comparisons across the audio and visual tracks.

\begin{center}
    \makebox[\textwidth][l]{\hspace*{0.07\textwidth}\includegraphics[width=1.08\textwidth]{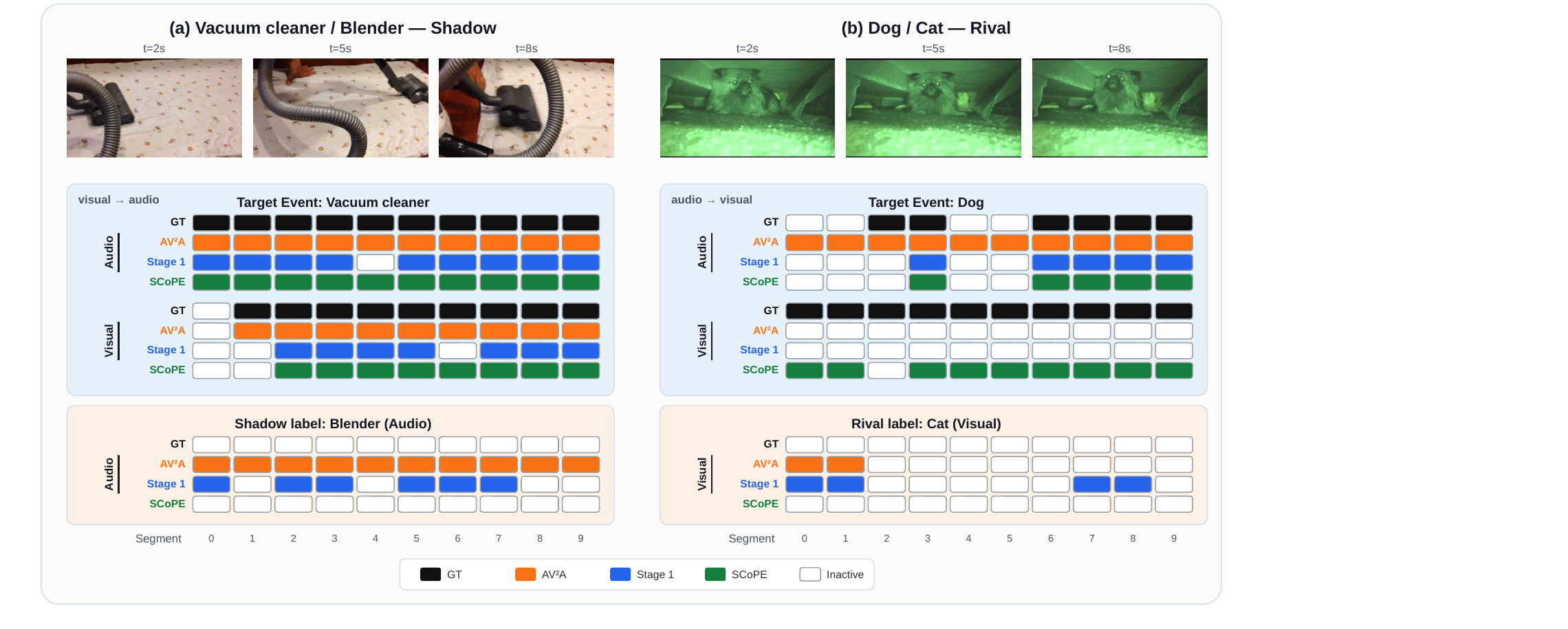}}
    \par\smallskip
    \refstepcounter{figure}
    \label{fig:qualitative_target_reselection}
    \parbox{\textwidth}{\small Figure~\thefigure: Detailed LLP cases. Blue cells show positive Stage~1 support, whereas green cells show the final Stage~2 prediction after re-solving and sparse readout. Upper blocks follow the annotated target event in both branches. Lower blocks follow a related candidate outside the annotation.}
\end{center}

\paragraph{How the target uses a source prior.}
The source modality does not send a binary decision. Equations~\eqref{eq:prior_strength}--\eqref{eq:stage2} pool its reconstruction-weighted coefficients, lower the target cost of source-supported events, and re-solve from the target embedding. A lower cost only lets the target reconsider an event; the target still needs enough residual evidence to assign it a positive coefficient. Equation~\eqref{eq:active_readout} then decides whether that coefficient enters the final prediction. Thus, below, ``accepted'' means retained by the final target prediction, not that SCoPE contains a separate binary gate for accepting a prior.

\paragraph{Vacuum cleaner and Blender.}
The first clip is annotated with Vacuum cleaner but not Blender. In the audio-to-visual direction, the audio source gives
$P_a(\text{Vacuum cleaner})=0.404$ and $P_a(\text{Blender})=0.113$.
After the fixed-budget normalization in Equation~\eqref{eq:weighted_cost}, their visual costs are $0.041$ and $0.272$, respectively, from the common base cost $\lambda_0=0.3$. Both visual Stage~1 coefficients are zero. After target re-selection, Vacuum cleaner has a positive visual coefficient in nine segments, ranging from $0.020$ to $0.084$, whereas every Blender coefficient remains zero. The visual evidence is therefore sufficient at the lower Vacuum cleaner cost but insufficient at the higher Blender cost. The source prior helps the visual branch recover Vacuum cleaner without inserting Blender.

The reverse visual-to-audio direction provides no class-specific guidance in this clip because the visual Stage~1 reconstruction confidence is zero in every segment, giving $P_v(c)=0$ for all events. The audio costs consequently remain at $\lambda_0$, and its Stage~1 and Stage~2 coefficient vectors are identical. The blue Blender cells in the lower block mark small positive Stage~1 coefficients; their absence from the green row is due to the final sparse readout, not a visual prior suppressing Blender.

\paragraph{Dog and Cat.}
The second clip is annotated with Dog but not Cat. The audio source gives both events a nonzero prior, but with different strengths: $P_a(\text{Dog})=0.417$ and $P_a(\text{Cat})=0.148$. Their visual costs become $0.124$ and $0.260$. Dog has no positive visual Stage~1 coefficient, yet re-selection produces coefficients between $0.016$ and $0.071$ wherever the visual residual supports it, and the final readout retains Dog across most of the clip. Cat is the stricter test because it also receives a nonzero prior. Its Stage~2 visual coefficients remain small, with nonzero values of $0.0002$, $0.0045$, and $0.0295$ in only three segments, and the largest-gap readout retains none of them. The weighted solve therefore considers both source-supported events, but target evidence and the common readout retain Dog and reject Cat. This is the precise sense in which the target controls the final decision.

\begin{center}
\begin{minipage}{.95\linewidth}
    \centering
    \includegraphics[width=\linewidth]{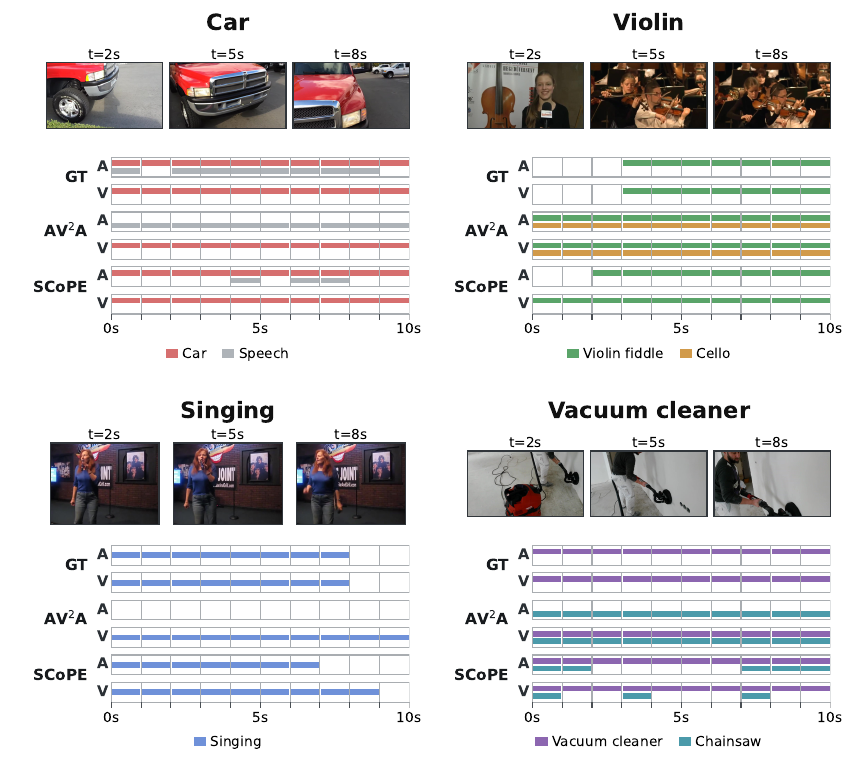}
    \par\smallskip
    \refstepcounter{figure}
    \label{fig:qualitative_more_examples}
    \parbox{\linewidth}{\small Figure~\thefigure: Additional qualitative LLP examples. Timelines compare selected ground-truth, AV$^2$A, and SCoPE audio and visual predictions. Each legend lists only the labels displayed in that panel.}
\end{minipage}
\end{center}

\subsection{Trained Reference Results}
\label{app:supervised_references}
Table~\ref{tab:supervised_references} places SCoPE beside trained systems only to show the remaining performance headroom. These rows are not controlled comparisons.

For LLP, the weakly supervised values reproduce Table~8 of the AV$^2$A~\cite{shaar2025av2a} supplementary material. The trained systems use task-specific networks and conventional AVVP features, while SCoPE performs inference with frozen CLIP+CLAP encoders.

For OV-AVEBench, we use the reported OV-AVEL values~\cite{zhou2025ovavel} and our reproduced SCoPE values.

\begin{center}
\begin{minipage}{\linewidth}
\centering
\refstepcounter{table}
\label{tab:supervised_references}
\small
\textbf{Table~\thetable:} Supervised reference results (\%). Rows are grouped by dataset and are not controlled comparisons.
\par\smallskip

\textbf{(a) LLP segment-level results}\par\smallskip
\setlength{\tabcolsep}{5.5pt}
\begin{tabular}{@{}lcccccc@{}}
\toprule
Method & Training & Audio & Visual & Audio-Visual & Type & Event \\
\midrule
MGN-MA~\cite{mo2022multimodal} & Weak & 60.3 & 55.3 & 50.1 & 55.3 & 56.9 \\
JoMoLD~\cite{cheng2022joint} & Weak & 61.1 & 63.5 & 56.8 & 60.5 & 59.7 \\
CMPAE~\cite{gao2023cmpae} & Weak & \textbf{64.1} & \textbf{66.1} & \textbf{59.1} & \textbf{63.3} & \textbf{62.9} \\
\midrule
\textbf{SCoPE} & Free & 40.32 & 59.06 & 54.37 & 51.25 & 39.74 \\
\bottomrule
\end{tabular}

\vspace{0.8em}
\textbf{(b) OV-AVEBench results}\par\smallskip
\setlength{\tabcolsep}{4.8pt}
\begin{tabular}{@{}lccccccc@{}}
\toprule
Method & Backbone & Training & Acc. & Seg-F1 & Event-F1 & Avg. & Unseen \\
\midrule
OV-AVEL~\cite{zhou2025ovavel} & ImageBind & Temporal & \textbf{67.1} & \textbf{56.9} & \textbf{49.5} & \textbf{57.8} & \textbf{55.8} \\
\textbf{SCoPE} & ImageBind & Free & 60.58 & 53.01 & 48.11 & 53.90 & 55.69 \\
\bottomrule
\end{tabular}
\end{minipage}
\end{center}

\end{document}